\documentclass[11pt,a4paper]{article}
\usepackage[hyperref]{acl2021}
\usepackage{times}
\usepackage{latexsym}

\usepackage{microtype}

\aclfinalcopy 

\usepackage{url}
\usepackage[T1]{fontenc}
\usepackage{amsmath}
\usepackage{amsfonts}
\usepackage{graphicx}
\usepackage{multirow, tabu}
\usepackage[title]{appendix}
\usepackage{arydshln}
\usepackage[acronyms]{glossaries}
\usepackage{glossaries-prefix}
\usepackage{comment}
\usepackage{soul}
\usepackage{enumitem}
\glsdisablehyper
\usepackage{arydshln}

\newcommand{\sectdot}[1]{Sec.~\ref{sec:#1}}

\newcommand{\ssectdot}[1]{Sec.~\ref{ssec:#1}}

\newcommand{\sssectdot}[1]{Sec.~\ref{sssec:#1}}

\newcommand{\eqndot}[1]{Eqn.~(\ref{eqn:#1})}
\newcommand{\fig}[1]{Figure~\ref{fig:#1}}
\newcommand{\figdot}[1]{Fig.~\ref{fig:#1}}
\newcommand{\tbl}[1]{Table~\ref{tab:#1}}

\newcommand{\twotbl}[2]{Tables~\ref{tab:#1} and \ref{tab:#2}}

\newcommand{\ignore}[1]{}

\makeatletter
\DeclareRobustCommand\onedot{\futurelet\@let@token\@onedot}
\def\@onedot{\ifx\@let@token.\else.\null\fi\xspace}

\def\eg{\emph{e.g}\onedot}

\definecolor{MyDarkBlue}{rgb}{0,0.08,1}
\definecolor{MyDarkGreen}{rgb}{0.02,0.6,0.02}
\definecolor{MyDarkRed}{rgb}{0.8,0.02,0.02}
\definecolor{MyDarkOrange}{rgb}{0.40,0.2,0.02}
\definecolor{MyPurple}{RGB}{111,0,255}
\definecolor{MyRed}{rgb}{1.0,0.0,0.0}
\definecolor{MyGold}{rgb}{0.75,0.6,0.12}
\definecolor{MyDarkgray}{rgb}{0.66, 0.66, 0.66}

\newacronym{ann}{ANN}{artificial neural network}
\newacronym{abus}{ABUS}{agenda-based user simulator}
\newacronym{nus}{NUS}{Neural User Simulator}
\newacronym{dst}{DST}{dialogue state tracking}
\newacronym{nlg}{NLG}{natural language generation}
\newacronym{rl}{RL}{reinforcement learning}
\newacronym{sl}{SL}{supervised learning}
\newacronym{pomdp}{POMDP}{partially observable markov decision proces}
\newacronym[prefixfirst={a\ }, prefix={an\ }]{us}{US}{user simulator}
\newacronym{ds}{DS}{dialogue system}
\usepackage{xspace}
\definecolor{green}{rgb}{0.0, 0.42, 0.24}
\setul{0.5ex}{0.3ex}
\setulcolor{red}

\title{Transferable Dialogue Systems and User Simulators}

\author{Bo-Hsiang Tseng$^{\dagger}$, Yinpei Dai$^{\ddagger}$, Florian Kreyssig$^{\dagger}$, Bill Byrne$^{\dagger}$ \\
${}^\dagger$Engineering Department, University of Cambridge, UK \\
${}^\ddagger$Alibaba Group \\
\texttt{\{bht26,flk24,wjb31\}@cam.ac.uk} \\
\texttt{yinpei.dyp@alibaba-inc.com}
}

\date{}

\begin{document}
\maketitle

\begin{abstract}

    One of the difficulties in training dialogue systems is the lack of training data. We explore the possibility of creating dialogue data through the interaction between a dialogue system and a user simulator. Our goal is to develop a modelling framework that can incorporate new dialogue scenarios through self-play between the two agents. In this framework, we first pre-train the two agents on a collection of source domain dialogues, which equips the agents to converse with each other via natural language. With further fine-tuning on a small amount of target domain data, the agents continue to interact with the aim of improving their behaviors using reinforcement learning with structured reward functions. In experiments on the MultiWOZ dataset, two practical transfer learning problems are investigated: 1) domain adaptation and 2) single-to-multiple domain transfer. We demonstrate that the proposed framework is highly effective in bootstrapping the performance of the two agents in transfer learning. We also show that our method leads to improvements in dialogue system performance on complete datasets.
\end{abstract}


\glsresetall
\section{Introduction}


This work aims to develop a modelling framework in which \glspl{ds} converse with \glspl{us} about complex topics using natural language. Although the idea of joint learning of two such agents has been proposed before, this paper is the first to successfully train both agents on complex multi-domain human-human dialogues and to demonstrate a capacity for transfer learning to low-resource scenarios without requiring re-redesign or re-training of the models.


One of the challenges in task-oriented dialogue modelling is to obtain adequate and relevant training data. A practical approach in moving to a new domain is via transfer learning, where pre-training on a general domain with rich data is first performed and then fine-tuning the model on the target domain. End-to-end \gls{ds} \cite{wenN2N17, li-etal-2017-end,dhingra-etal-2017-towards} are particularly suitable for transfer learning, in that such models are optimised as a single system. By comparison, pipe-lined based \glspl{ds} with multiple individual components \cite{POMDP} require fine-tuning of each component system. These separate steps can be done independently, but it becomes difficult to ensure optimality of the overall system.

A similar problem arises in the data-driven \gls{us} as commonly used in interaction with the \gls{ds}. Though many \glspl{us} have been proposed and been widely studied, they usually operate at the level of semantic representation \cite{Kreyssig2018NeuralUS,Asri2016ASM}. These models can capture user intent, but are otherwise somewhat artificial as user simulators in that they do not consume and produce natural language. As discussed above for \glspl{ds}, the end-to-end architecture for the \gls{us} also offers simplicity in transfer learning across domains.

There are also potential advantages to continued joint training of the \gls{ds} and the \gls{us}. If a user model is less than perfectly optimised after supervised learning over a fixed training corpus, further learning through interaction between the two agents offers the \gls{us} the opportunity to refine its behavior. Prior work has shown benefits from this approach to dialogue policy learning, with a higher success rate at dialogue level \cite{Liu2017IterativePL,papangelis-etal-2019-collaborative,takanobu2020multi}, but there has not been previous work that addresses multi-domain end-to-end dialogue modelling for both agents. \citet{takanobu2020multi} address refinement of the dialogue policy alone at the semantic level, but do not address end-to-end system architectures. \citet{Liu2017IterativePL,papangelis-etal-2019-collaborative} address single-domain dialogues \cite{henderson-etal-2014-second}, but not the more realistic and complex multi-domain dialogues.

This paper proposes a novel learning framework for developing dialogue systems that performs \textbf{J}oint \textbf{O}ptimisation with a \textbf{U}ser \textbf{S}imula\textbf{T}or (\textbf{JOUST}).\footnote{The code is released at \url{https://github.com/andy194673/joust}.}
Through the pre-training on complex multi-domain datasets, two agents are able to interact using natural language, and further create more diverse and rich dialogues. Using \gls{rl} to optimise both agents enables them to depart from known strategies learned from a fixed limited corpus, to explore new, potentially better policies. Importantly, the end-to-end designs in the framework makes it easier for transfer learning of two agents from one domain to another. We also investigate and compare two reward designs within this framework: 1) the common choice of task success at dialogue level; 2) a fine-grained reward that operates at turn level.
Results on  MultiWOZ dataset \cite{budzianowski2018multiwoz} show that our method is effective in boosting the performance of the \gls{ds} in complicated multi-domain conversation.
To further test our method in more realistic scenarios, we design specific experiments on two low-resource setups that address different aspects of data sparsity.
Our contributions can be summarised as follows:

\begin{itemize}
    \item Novel contributions in joint optimisation of a fully text-to-text dialogue system with a matched user simulator on complex, multi-domain human-human dialogues.
    \item Extensive experiments, including exploring different types of reward, showing that our framework with a learnable \gls{us} boost overall performance and reach new state-of-the-art performance on MultiWOZ.
    \item Demonstration that our framework is effective in two transfer learning tasks of practical benefit in low-resources scenarios with in-depth analysis of the source of improvements.
\end{itemize}

\begin{figure*}[t!]
    \centering
    \includegraphics[width=\linewidth]{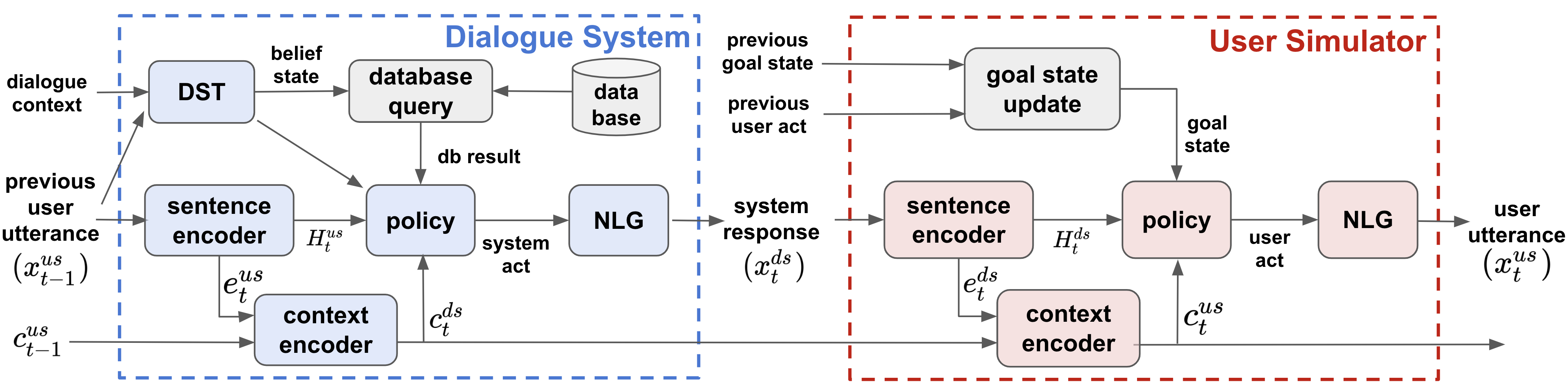}
    \caption{Overall architecture of the proposed framework, where \glsreset{ds}\glsreset{us}the \gls{ds} and \gls{us} discourse with each other. $t$ denotes dialogue turn index. The context encoder is shared between the two agents.}
    \label{fig:concate_model}
\end{figure*}
\section{Pre-training the Dialogue System and User Simulator}\label{sec:SL}
In our joint learning framework, we first pre-train the \gls{ds} and \gls{us} using supervised learning so that two models are able to interact via natural language. This section presents the architectures of two agents, illustrated in \figdot{concate_model}, and the objectives used for supervised learning.

\subsection{Dialogue system}\label{ssec:dialoguesystem}
\paragraph{\texorpdfstring{\Gls{dst}}{Dialogue state tracking}}
The first task of a \gls{ds} is to process the dialogue history in order to maintain the belief state which records essential information of the dialogue. A \gls{dst} model is utilized to predict the set of slot-value pairs which constitute the constraints of the entity for which the user is looking for, \eg \{\texttt{hotel\_area}=\texttt{north}, \texttt{hotel\_name}=\texttt{gonville\_hotel}\}.

The \gls{dst} model used here is an encoder-decoder model with attention mechanism \cite{Bahdanau2015NeuralMT}. The set of slot-value pairs is formulated as a slot sequence together with a value sequence.
For the $t^\text{th}$ dialogue turn, the \gls{dst} model first encodes the dialogue context and the most recent user utterance $x^{us}_{t-1}$ using a bi-directional LSTM \cite{HochSchm97} to obtain hidden states $H_t^{enc}=\{h^{enc}_{1},...,h^{enc}_{j},...\}$.
At the $i^\text{th}$ decoding step of turn $t$, the previous decoder hidden state $h^{dec}_{i-1}$ is used to attend over $H_t^{enc}$ to obtain the attention vector $a_{i}$.
The decoder takes $a_{i}$, $h^{dec}_{i-1}$ and the embedding of the slot token predicted at $i-1$ to produce the current hidden state $h^{dec}_{i}$.
The $h^{dec}_{i}$ is then passed through separate affine transforms followed by the softmax function to predict a slot token and value for step $i$. The final belief state is the aggregation of predicted slot-value pairs of all decoding steps.

\paragraph{Database Query}
Based on the updated belief state, the system searches the database and retrieves the matched entities.
In addition, a one-hot vector of size 3 characterises the result of every query.

\paragraph{Context Encoding}
To capture the dialogue flow, a hierarchical LSTM \cite{serban2016building} encodes the dialogue context from turn to turn throughout the dialogue. At each turn $t$, the most recent user utterance $x^{us}_{t-1}$ is encoded by an LSTM-based sentence encoder to obtain a sentence embedding $e^{us}_{t}$ and hidden states $H^{us}_{t}$. Another LSTM is used as the context encoder, which encodes $e^{us}_{t}$ as well as the output of the context encoder on the user side $c^{us}_{t-1}$ from the previous turn (see \figdot{concate_model}). The context encoder produces the next dialogue context state $c^{ds}_{t}$ for the downstream dialogue manager.

\paragraph{Policy}
The dialogue manager determines the system dialogue act based on the current state of the dialogue. 
The system dialogue act is treated as a sequence of tokens in order to handle cases in which multiple system actions exist in the same turn.
The problem is therefore formulated as a sequence generation task using an LSTM. At each decoding step, the inputs to the policy decoder are: 1) the embedding of the act token predicted at the previous step; 2) the previous hidden state; 3) the attention vector obtained by attending over the hidden states of the user utterance $H^{us}_{t}$ using 2) as query; 4) the database retrieval vector; 5) the summarized belief state, which is a binary vector where each entry corresponds to a domain-slot pair.
The output space contains all possible act tokens. For better modeling of the dialogue flow, the initialization of the hidden state is set to the context state $c^{ds}_{t}$ obtained by the context encoder.

\paragraph{\texorpdfstring{\Gls{nlg}}{Natural language generation}}
The final task of the \gls{ds} is to generate the system response, based on the predicted system dialogue act. 
To generate the word sequence another LSTM is used as the NLG model. 
At each decoding step, the previous hidden state serves as a query to attend over the hidden states of the policy decoder.
The resulting attention vector and the embedding of the previous output word are the inputs to an LSTM whose output is the word sequence with {\em delexicalized} tokens.
These {\em delexicalized} tokens will be replaced by retrieval results to form the final utterance.

\subsection{User Simulator}\label{ssec:usersim}
As in the \gls{ds}, the proposed \gls{us} has a dialogue manager, an \gls{nlg} model and a dialogue context encoder. However, in place of a \gls{dst} to maintain the belief state, the \gls{us} maintains an internal goal state to track progress towards satisfying the user goals.

\paragraph{Goal State}
The goal state is modelled as a binary vector that summarises the dialogue goal. Each entry of the vector corresponds to a domain-slot pair in the ontology. At the beginning of a dialogue, goal state entries are turned on for all slots that make up the goal.
At each dialogue turn, the goal state is updated based on the previous user dialogue act.
If a slot appears in the previous dialogue act, either as information from the user or as a request by the \gls{us}, the corresponding entry is turned off.

\paragraph{Context encoding, Policy \& NLG in the \gls{us}}
These steps follow their implementations in the \gls{ds}. For context encoding in the \gls{us}, a sentence encoder first encodes the system response using an LSTM to obtain hidden states $H^{ds}_{t}$ and sentence embedding $e^{ds}_{t}$. The context encoder takes $e^{ds}_{t}$ and \gls{ds} context state $c^{ds}_{t}$ as inputs to produce the dialogue context state $c^{us}_{t}$ which is passed to the \gls{ds} at the next turn.

Also as in the \gls{ds}, the policy and the \gls{nlg} model of the \gls{us} are based on LSTMs.
The input to the policy are goal state, hidden states of the sentence encoder $H^{ds}_{t}$ and context state $c^{us}_{t}$, to produce the user dialogue act, represented as in the \gls{ds} as a sequence of tokens.
The \gls{nlg} model takes the hidden states of policy decoder as input to generate the user utterance, which is then {\em lexicalised} by replacing {\em delexicalised} tokens using the user goal.

\subsection{Supervised Learning}\label{ssec:SLOptimization}
For each dialogue turn, the ground truth dialogue acts and the output word sequences are used as supervision for both the \gls{ds} and the \gls{us}. The losses of the policy and the \gls{nlg} model are the cross-entropy losses of the predicted sequence probability $p$ and the ground-truth $y$:
\begin{equation}
\begin{aligned}
    & L^{*}_{pol} = \sum^{|A|}_{i=1} - y^{*}_{a,i} \log p^{*}_{a,i} \\
    & L^{*}_{nlg} = \sum^{|W|}_{i=1} - y^{*}_{w,i}  \log p^{*}_{w,i}
\end{aligned}
\end{equation}
In the above, * can be either $ds$ or $us$, referring either to the \gls{ds} or the \gls{us}: e.g. $p^{ds}_{a, i}$ is the probability of the system act token at the $i^{\text{th}}$ decoding step in a given turn.
The ground-truth $y$ contains both word sequences and act sequences with $W$ and $A$ as their lengths.

The \gls{dst} annotations are also used as supervision for the \gls{ds}.
The loss of the \gls{dst} model is defined as the sum of the cross-entropy losses for slot and value:
\begin{equation}
\label{eq:loss_dst}
    L^{ds}_{dst} = \sum^{|SV|}_{i=1} - y^{ds}_{s,i} \log p^{ds}_{s,i} -  y^{ds}_{v,i} \log p^{ds}_{v,i}
\end{equation}
where $|SV|$ is the number of slot-value pairs in a turn; $i$ is the decoding step index. $p^{ds}_{s,i}$ and $p^{ds}_{v,i}$ are the predictions of slot and value at the $i^{\text{th}}$ step.

The overall losses for the \gls{ds} and the \gls{us} are:
\begin{equation}
\begin{aligned}
    & L^{ds}(\theta^{ds}) = L^{ds}_{dst} + L^{ds}_{pol} + L^{ds}_{nlg} \\
    & L^{us}(\theta^{us}) = L^{us}_{pol} + L^{us}_{nlg} 
\end{aligned}
\end{equation}
where $\theta^{ds}$ and $\theta^{us}$ are the parameters of \gls{ds} and \gls{us}, respectively.
The two agents are updated jointly to minimize the sum of the losses ($L^{ds}\!+\! L^{us}$).
The success rate of the generated dialogues is used as the stopping criterion for supervised learning.
\section{RL Optimisation of the Dialogue System and User Simulator}\label{sec:RL}
After the DS and US models are pre-trained from the corpus using supervised learning, they are fine-tuned using reinforcement learning (\gls{rl}) based on the dialogues generated during their interactions.
Two reward designs are presented after which the optimisation strategy is given.

\subsection{Dialogue-Level Reward}\label{sec:dial_r}
Following common practice \cite{el2014task, su-etal-2017-sample, casanueva2018feudal, zhao2019rethinking}, the success of the simulated dialogues is used as the reward, which can only be observed at the end of the dialogue.
A small penalty is given at each turn to discourage lengthy dialogues.
When updating the \gls{us} jointly with the \gls{ds} during interaction using RL, the reward is shared between two agents.


\subsection{Turn-Level Reward}\label{sec:turn_r}
While the dialogue-level reward is straight-forward, it only considers the final task success rate of the dialogues and neglects the quality of the individual turns.
For complex multi-domain dialogues there is a risk that this will make it difficult for the system to learn the relationship between actions and rewards.
We thus propose a turn-level reward function that encapsulates the desired behavioural features of fundamental dialogue tasks.
The rewards are designed separately for the \gls{us} and the \gls{ds} according to their characteristics. 

\paragraph{\gls{ds} Reward} A good \gls{ds} should learn to refine the search by requesting needs from the user and providing the correct entities, with their attributes, that the user wishes to know. Therefore at the current turn a positive reward is assigned to \gls{ds} if: 1) it requests slots that it has not requested before; 2) it successfully provides an entity; or 3) is answers correctly all additional attributes requested by the user. Otherwise, a negative reward is given.

\paragraph{\gls{us} Reward} A good \gls{us} should not repeatedly give the same information or request attributes that have already been provided by the \gls{ds}. Therefore, a positive reward is assigned to the \gls{us} if: 1) it provides new information about slots; 2) it asks new attributes about a certain entity, or 3) it replies correctly to a request from the \gls{ds}. Otherwise a penalty is given.


\subsection{Optimization}
We apply the \emph{Policy Gradient Theorem}~\cite{NIPS1999_1713} to the space of (user/system) dialogue acts.
In the $t^{\text{th}}$ dialogue turn, the reward $r^{ds}_{t}$ or $r^{us}_{t}$ is assigned to the two agents at final last step of their generated act sequence.
The return for the action at the $i^{\text{th}}$ step is $R^{*}_{i}=\gamma^{|A^{*}|-i} r^{*}_{t}$,
where $*$ denotes $ds$ or $us$, and $|A^*|$ is the length of the act sequence of each agent. $\gamma\!\in\![0,1]$ is a discounting factor. The policy gradient of each turn can then be written as:
\begin{equation}
    \label{eqn:policy_gradient}
    \nabla_{\theta^*} J^{*}(\theta^{*}) =  \sum^{|A^{*}|}_{i} R^{*}_{i} \nabla_{\theta^*}\log p^{*}_{a, i}
\end{equation}
where $p^{*}_{a,i}$ is the probability of the act token at the $i^{\text{th}}$ step in the predicted dialogue act sequence.
The two agents are updated using \eqndot{policy_gradient} at each turn within the entire simulated dialogue.
\section{Experiments}

\paragraph{Dataset}
The MultiWOZ 2.0 dataset \cite{budzianowski2018multiwoz} is used for all experiments.
It contains 10.4k dialogues with an average of 13.6 turns.
Each dialogue can span up to three domains.
Compared to previous benchmark corpora such as DSTC2 \cite{Williams2016TheDS} or WOZ2.0 \cite{wenN2N17}, MultiWOZ is more challenging because 1) its rich ontology contains 39 slots across 7 domains; 2) the \gls{ds} can take multiple actions in a single turn; 3) the complex dialogue flow makes it difficult to hand-craft a rule-based \gls{ds} or an agenda-based \gls{us}. \citet{lee-etal-2019-convlab} provided the user act labels.

\paragraph{Training Details}
The positive and negative \gls{rl} rewards of \sectdot{RL} are tuned in the range [-5, 5] based on the dev set.
The user goals employed for interaction during \gls{rl} are taken from the training data without synthesizing new goals.
Further training details can be found in Appendix A.1.

\paragraph{Evaluation Metrics}
The proposed model is evaluated in terms of the inform rate (Info), the success rate (Succ), and BLEU.\footnote{For a fair comparison to previously proposed models, the same evaluation script provided by the MultiWOZ organizers \url{https://github.com/budzianowski/multiwoz} is used and the official data split for train/dev/test is followed.} 
The inform rate measures whether the \gls{ds} provides the correct entity matching the user goal, while the success rate further requires the system to answer all user questions correctly.
Following \cite{mehri-etal-2019-structured}, the combined performance (Comb) is also reported, calculated as $0.5*(\text{Info}+\text{Succ}) + \text{BLEU}$.

\subsection{Interaction Quality} \label{ssec:result_with_us}
First, it is examined whether the proposed learning framework improves the discourse between dialogue system and user simulator. Several variants of our model are examined: 1) two agents are pre-trained using supervised learning, serving as baseline; 2) RL is used to fine-tune only the DS (RL-DS) or both agents (RL-Joint). In each RL case, we can either use rewards at the dialogue level (dial-R, \sectdot{dial_r}) or rewards at the turn-level (turn-R, \sectdot{turn_r}). The two trained agents interact based on 1k user goals from the test corpus, with the generated dialogues being evaluated using the metrics above.

From \tbl{interact}, we can see that the application of RL in our framework improves the success rate by more than 10\% (b-e vs. a).
This indicates that the \gls{ds} learns through interaction with the learned \gls{us}, and the designed rewards, to be better at completing the task successfully.
Moreover, the joint optimisation of both the \gls{us} and the \gls{ds} provides dialogues with higher success rate than only optimising the \gls{ds} (c\&e vs. b\&d).
It shows that the behaviour of the \gls{us} is realistic enough and diverse enough to interact with the \gls{ds}, and its behavior can be improved together during RL optimisation.
Finally, by comparing two reward designs, the fine-grained rewards at the turn level seem to be more effective towards guiding two agents' interaction (b\&c vs. d\&e), which is reasonable since they reflect more than simple success rate in terms of the nature of the tasks.
Some real, generated dialogues through the interactions are provided in Appendix A.6; we note that after \gls{rl}, both agents respond to requests more correctly and also learn not to repeat the same information, leading to a more successful and smooth interaction without loops in the dialogue.
The corresponding error analysis of each of the agents is provided later in \sectdot{error_analysis}.


\begin{table}[t]
\centering
\resizebox{0.75\linewidth}{!}{%
\begin{tabu}{lcc}
\tabucline [1pt]{1}
Model & Info & Succ \\ \hline
(a) Supervised Learning & 69.77 & 58.02  \\ \hdashline
(b) RL-DS w/ dial-R & 81.38 & 70.67 \\
(c) RL-Joint w/ dial-R & 82.83 & 71.57  \\ \hdashline
(d) RL-DS w/ turn-R & 85.62  & 70.34 \\
(e) RL-Joint w/ turn-R & \textbf{86.49} & \textbf{73.04} \\
\tabucline [1pt]{1}
\end{tabu}%
}
\caption{Quality for dialogues generated by two agents in JOUST using the test corpus user goals. BLEU is not reported since no reference sentences are available for these interactions.}
\label{tab:interact}
\end{table}

\begin{table}[t]
\centering
\resizebox{1.\linewidth}{!}{%
\begin{tabu}{lcccc}
\tabucline [1pt]{1}
\multicolumn{1}{c}{Model} & Info & Succ & BLEU & Comb \\ \tabucline [1pt]{1}
HRED-TS \cite{peng2019teacher} & 70.0 & 58.0 & 17.5 & 81.5 \\
DAMD \cite{Zhang2019TaskOrientedDS} & 76.3 & 60.4 & 16.6 & 85.0 \\ 
SimpleTOD$^{*}$\cite{hosseini2020simple} & 84.4  & 70.1 & 15.0 & 92.3 \\
SOLOIST$^{*}$ \cite{peng2020soloist} & \textbf{85.5} & 72.9 & 16.5 & 95.7 \\
MinTL-BART$^{*}$ \cite{lin-etal-2020-mintl} & 84.9 & \textbf{74.9} & \textbf{17.9} & \textbf{97.8}  \\ \hdashline
JOUST Supervised Learning & 77.4  & 66.7 & 17.4 & 89.5 \\
JOUST RL-Joint w/ dial-R & 80.6 & 69.4 & 17.5 & 92.5 \\
JOUST RL-Joint w/ turn-R & \textbf{83.2} & \textbf{73.5} & \textbf{17.6} & \textbf{96.0} \\
\tabucline [1pt]{1}
\end{tabu}%
}
\caption{Empirical comparison with state-of-the-art dialogue systems using the predicted belief state. $^{*}$ indicates leveraging of pre-trained transfomer-based models.}
\label{tab:main-dst}
\end{table}

\subsection{Benchmark Results} \label{ssec:benchmark}
We conduct experiments on the official test set for comparison to existing end-to-end \glspl{ds}.
The trained \gls{ds} is used to interact with the fixed test corpus following the same setup of \citet{budzianowski2018multiwoz}. Results are reported using a predicted belief state (\tbl{main-dst}) and using an oracle belief state (\tbl{main-oracle}).
In general, we can observe similar performance trends as in \ssectdot{result_with_us} with RL optimization of our model.
Joint learning of two agents using RL with the fine-grained rewards reaches the best combined score and success rate.
This implies that the exploration of more dialogue states and actions in the simulated interactions reinforces the behaviors that lead to higher success rate, and that these generalise well to unfamiliar states encountered in the test corpus.

Our best RL model produces competitive results in \tbl{main-dst} when using predicted belief state, and can further outperform the previous work in \tbl{main-oracle} when using oracle belief state.
Note that we do not leverage the powerful pre-trained transformer-based models like SOLOIST or MinTL-BART model.
We found that with RL optimisation, our LSTM-based models can still perform competitively.
In terms of \gls{ds} model structure, the most similar work would be the DAMD model.
The performance gain found in comparing "JOUST Supervised Learning" to DAMD is partially due to the better performance of our DST model.\footnote{In correspondence, the DAMD authors report a DST model with joint accuracy of ca. 35\%, while ours is 45\%.}

We also conduct experiments using only 50\% of the training data for supervised learning to verify the efficacy of the proposed method under different amounts of data.
As shown in \tbl{50sota}, it is observed that our method also improves the model upon supervised learning when trained with less data and the improvements are consistent with the complete data scenario.

\begin{table}[t]
\centering
\resizebox{1.\linewidth}{!}{%
\begin{tabu}{lcccc}
\tabucline [1pt]{1}
\multicolumn{1}{c}{Model} & Info  & Succ & BLEU & Comb \\ \tabucline [1pt]{1}
SimpleTOD$^{*}$\cite{hosseini2020simple} & 88.9  & 67.1 & 16.9 & 94.9 \\
MoGNet \cite{pei2019retrospective} & 85.3 & 73.3 & 20.1 & 99.4  \\
ARDM$^{*}$ \cite{wu2019alternating} & 87.4 & 72.8 & 20.6 & 100.7 \\
DAMD \cite{Zhang2019TaskOrientedDS} & 89.2 & 77.9 & 18.6 & 102.2 \\
SOLOIST$^{*}$ \cite{peng2020soloist} & 89.6 & \textbf{79.3} & 18.3 & 102.5 \\ 
PARG \cite{Gao2020ParaphraseAT} & 91.1 & 78.9 & 18.8 & 103.8 \\
MarCo$^{*}$ \cite{wang2020multi} & \textbf{92.3} & 78.6 & \textbf{20.0} & \textbf{105.5} \\
\hdashline
JOUST Supervised Learning & 88.5 & 79.4 & 18.3 & 102.3 \\
JOUST RL-Joint w/ dial-R & 93.9 & 85.7 & 16.9 & 106.7 \\
JOUST RL-Joint w/ turn-R & \textbf{94.7} & \textbf{86.7} & \textbf{18.7} & \textbf{109.4} \\
\tabucline [1pt]{1}
\end{tabu}%
}
\caption{Empirical comparison with state-of-the-art dialogue systems using oracle belief state. $^{*}$ indicates leveraging of pre-trained transfomer-based models.}
\label{tab:main-oracle}
\end{table}

\begin{table}[t]
\centering
\resizebox{0.9\linewidth}{!}{%
\begin{tabu}{lcccc}
\tabucline [1pt]{1}
Model & Info. & Succ. & BLEU & Comb. \\ \hline
\multicolumn{5}{c}{Belief State = Predicted} \\ \hline
Supervised Learning & 70.37 & 55.43 & 17.29 & 80.19 \\
RL-Joint w/ turn-R & \textbf{74.83} & \textbf{60.60} & \textbf{17.41} & \textbf{85.12} \\ \tabucline [1pt]{1}
\multicolumn{5}{c}{Belief State = Oracle} \\ \hline
Supervised Learning & 89.67 & 74.5 & 16.96 & 99.04 \\
RL-Joint w/ turn-R & \textbf{94.27} & \textbf{81.47} & \textbf{17.20} & \textbf{105.06} \\
\tabucline [1pt]{1}
\end{tabu}%
}
\caption{Results of JOUST using 50\% training data in supervised learning.}
\label{tab:50sota}
\end{table}

\subsection{Transfer Learning}\label{ssec:transferlearning}
In this section, we demonstrate the capability of transfer learning of the proposed framework under two low-resource setups: Domain Adaptation and Single-to-Multiple Domain Transfer.
Two fine-tuning methods are adopted: the straightforward fine-tuning without any constraints (Naive) and elastic weight consolidation (EWC) \cite{kirkpatrick2017overcoming}.
We show that the proposed RL can be further applied to both methods and produces significantly improved results. Here we experiment the best RL variants using turn-level rewards (same as (e) in \tbl{interact}).


\paragraph{Domain Adaptation}
In these experiments, each of five domains is selected as the target domain.
Taking the \textit{hotel} domain for example, 300 dialogues\footnote{For each domain, 300 dialogues accounts for 10\% of all target-domain data. Refer to Appendix A.2 for data statistics.} involving the hotel domain are sampled from the training corpus as adaptation data. The rest of the dialogues, not involving the hotel domain, form the source data.
Both the \gls{ds} and the \gls{us} are first trained on the source data (Source), and then fine-tuned on the limited data of the target domain (Naive, EWC). Afterwards, the pair of agents is trained in interaction using the proposed \gls{rl} training regime (+RL).

Results in the form of the combined score are given in \tbl{domain_transfer} (corresponding success rates are provided in Appendix A.5). As expected, models pre-trained on source domains obtain low combined scores on target domains. Fine-tuning using Naive or EWC method significantly bootstraps the systems, where the regularization in EWC benefits more for the low-resource training.
By applying our proposed framework to the two sets of fine-tuned models, the performance can be further improved by 7-10\% in averaged numbers, with both predicted and oracle belief states.
This indicates that through the interaction with the \gls{us}, the \gls{ds} is not constrained by having seen only a very limited amount of target domain data, and that it can learn effectively from the simulated dialogues using the simple reward structure (the RL learning curve is presented in \sssectdot{curve}).
With a better initialization points such as EWC models, the models can learn from a higher quality interaction and produce better results (EWC+RL vs Naive+RL).
On average, the final performance obtained by EWC+RL model doubles that of Source model, which demonstrates the efficacy of the proposed method in domain adaptation.


\begin{table}[tb]
\centering
\resizebox{\linewidth}{!}{%
\begin{tabu}{lccccc||c}
\tabucline [1pt]{1}
\textbf{Model} & \textbf{Restaurant} & \textbf{Hotel} & \textbf{Attraction} & \textbf{Train} & \textbf{Taxi} & \textbf{Avg.} \\ \hline
\multicolumn{7}{c}{Belief State = Predicted} \\ \hline
Source & 21.1 & 28.6 & 25.2 & 59.6 & 48.7 & 36.6 \\ \hdashline
Naive & 46.7 & 56.2 & 66.1 & 68.5 & 66.3 & 60.8 \\
EWC & 56.7 & 58.2 & 71.6 & 69.3 & 78.7 & 66.9 \\ \hdashline
Naive+RL & 57.0 & 66.8 & 72.5 & \textbf{72.3} & 75.4 & 68.8 \\
EWC+RL & \textbf{64.6} & \textbf{67.8} & \textbf{75.8} & 71.6 & \textbf{87.6} & \textbf{73.5} \\ \tabucline [1pt]{1}
\multicolumn{7}{c}{Belief State = Oracle} \\ \hline
Source & 33.2 & 40.1 & 34.3 & 70.7 & 55.4 & 46.7 \\ \hdashline
Naive & 85.6 & 84.2 & 77.9 & 96.7 & 93.4 & 87.5 \\
EWC & 84.1 & 85.1 & 89.8 & 101.7 & 97.5 & 91.6 \\ \hdashline
Naive+RL & \textbf{97.6} & 99.2 & 88.5 & 104.0 & 103.4 & 98.5 \\
EWC+RL & 97.5 & \textbf{100.7} & \textbf{96.0} & \textbf{104.9} & \textbf{106.3} & \textbf{101.1} \\
\tabucline [1pt]{1}
\end{tabu}%
}
\caption{Combined scores in domain adaptation. 300 dialogues are used for each target domain adaptation.}
\label{tab:domain_transfer}
\end{table}


\paragraph{Single-to-Multiple Domain Transfer}
Another transfer learning scenario is investigated where only limited multi-domain data is accessible but sufficient single-domain dialogues are available. This setup is based on a practical fact that single-domain dialogues are often easier to collect than multi-domain ones. All single-domain dialogues in the training set form the source data.
For each target multi-domain combination, 100 dialogues\footnote{There are 6 types of domain combinations in MultiWOZ, as shown in \tbl{single-to-multi}. For each multi-domain combination, 100 dialogues accounts for 11\% of its multi-domain data.} are sampled as adaptation data.
As before, the \gls{ds} and the \gls{us} are first pre-trained on the source data and then fine-tuned on the adaptation data. Afterwards, two agents improve themselves through interaction. The models are tested using the multi-domain dialogues of the test corpus.


Results in the form of the combined score are given in Table \ref{tab:single-to-multi} (refer to Appendix A.5 for success rates).
Although the Source models capture individual domains, they cannot manage the complex flow of multi-domain dialogues and hence produce poor combined scores, with worst results on combinations of three domains.
Fine-tuning improves performance significantly, as the systems learn to transition between domains in the multi-domain dialogue flow.
Finally, applying our RL optimization further increases the performance by 6-9\% on average.
This indicates that the dialogue agents can learn more complicated policies through exploring more dialogue states and actions while interacting with user simulator.
We analyse the sources of improvements in the following section.


\begin{table}[tb]
\centering
\resizebox{\linewidth}{!}{%
\begin{tabu}{lcccccc||c}
\tabucline [1pt]{1}
\textbf{Model} & \multicolumn{1}{l}{\textbf{H+T}} & \multicolumn{1}{l}{\textbf{R+T}} & \multicolumn{1}{l}{\textbf{A+T}} & \multicolumn{1}{l}{\textbf{A+H+X}} & \multicolumn{1}{l}{\textbf{H+R+X}} & \multicolumn{1}{l}{\textbf{A+R+X}} & \textbf{Avg.} \\ \tabucline [1pt]{1}
\multicolumn{8}{c}{Belief State = Predicted} \\ \hline
Source & 46.0 & 55.4 & 34.3 & 22.0 & 26.6 & 19.9 & 34.0 \\ \hdashline
Naive & 57.2 & 69.2 & 65.0 & 40.3 & 36.0 & 42.8 & 51.7 \\
EWC & 57.4 & 72.1 & 66.1 & 43.7 & 39.0 & 45.0 & 53.9 \\ \hdashline
Naive+RL & 63.2 & 74.4 & \textbf{68.4} & \textbf{47.4} & 42.7 & \textbf{48.7} & 57.5 \\
EWC+RL & \textbf{64.7} & \textbf{77.6} & 67.6 & 46.6 & \textbf{43.2} & 48.5 & \textbf{58.0} \\ \tabucline [1pt]{1}
\multicolumn{8}{c}{Belief State = Oracle} \\ \hline
Source & 82.3 & 93.3 & 76.2 & 36.8 & 55.4 & 42.4 & 64.4 \\ \hdashline
Naive & 88.8 & 98.4 & 85.9 & 72.2 & 79.8 & 76.7 & 83.6 \\
EWC & 95.5 & 96.9 & 89.6 & 70.0 & 81.5 & 79.6 & 85.5 \\ \hdashline
Naive+RL & 99.7 & \textbf{104.3} & 92.0 & 80.6 & \textbf{97.2} & \textbf{89.3} & 93.9 \\
EWC+RL & \textbf{100.2} & 103.0 & \textbf{93.9} & \textbf{82.6} & 95.0 & 89.2 & \textbf{94.0} \\
\tabucline [1pt]{1}
\end{tabu}%
}
\caption{Combined scores in single-to-multiple domain transfer where 100 dialogues on each target scenario are used for adaptation. R, H, A, T, X represent Restaurant, Hotel, Attraction, Train, Taxi domain.}
\label{tab:single-to-multi}
\end{table}

\subsection{Analysis} \label{ssec:analysis}

\subsubsection{Error Analysis}\label{sec:error_analysis}
We first examine the behavior of the \gls{us} and the \gls{ds} to understand the improved success rate in transfer learning.
The models are those of \tbl{domain_transfer} and are examined after fine-tuning using Naive method (Naive) and then after reinforcement learning (Naive+RL).
For the \gls{ds}, the rates of missing entities (Miss Ent.) and of wrong answers (Wrong Ans.) are reported.
For the \gls{us}, rates of repetitions of attributes (Rep. Att.) and of missing answers (Miss Ans.) are reported.
The results shown in \tbl{error} are averaged over the five adaptation domains\footnote{Results for each domain can be found in Appendix A.3.}.
We see that with RL optimisation the errors made by the two agents are reduced significantly.
Notably, the user model learns not to repeat the information already provided and attempts to answer more of the questions from the dialogue agent.
These are the behaviors the reward structure of \sectdot{turn_r} are intended to encourage, and they lead to more successful interactions in policy learning.


\begin{table}[t]
\centering
\resizebox{0.975\linewidth}{!}{%
\begin{tabu}{l|cc||cc}
\tabucline [1pt]{1}
\multicolumn{1}{c|}{\multirow{2}{*}{\textbf{Model}}} & \multicolumn{2}{c||}{\textbf{Dialogue System}} & \multicolumn{2}{c}{\textbf{User Simulator}} \\
\multicolumn{1}{c|}{} & Miss Ent. & Wrong Ans. & Rep. Att. & Miss Ans. \\ \tabucline [1pt]{1}
Naive & 17.59 & 36.99 & 10.12 & 47.27 \\
Naive+RL & \textbf{2.73} & \textbf{9.54} & \textbf{1.47} & \textbf{32.60} \\
\tabucline [1pt]{1}
\end{tabu}%
}
\caption{Error analysis (\%) of the \gls{us} and the \gls{ds} agents averaged over 5 adaptation domains. Lower is better.}
\label{tab:error}
\end{table}

\subsubsection{Exploration of States and Actions}
We now investigate whether our framework encourages exploration through increased interaction in transfer learning.
We report the number of unique belief states in the training corpus and in the dialogues generated during \gls{rl} interaction, as well as the unique action sequences per state that each agent predicts.

As shown in \tbl{exploration}, the DS encounters more states in interaction with the US and also takes more unique actions in reinforcement learning relative to what it sees in supervised learning.
In this way the \gls{ds} considers additional strategies during the simulated training dialogues, with the opportunity to reach better performance even with only limited supervised data.
Detailed results for each adaptation case are provided in Appendix A.4.

\begin{table}[t]
\centering
\resizebox{1\linewidth}{!}{%
\begin{tabu}{l|cc||cc}
\tabucline [1pt]{1}
\multicolumn{1}{c|}{} & \multicolumn{2}{c||}{\textbf{Domain adaptation}} & \multicolumn{2}{c}{\textbf{Single-to-Multiple}} \\ 
\multicolumn{1}{c|}{} & states & actions & states & actions \\ \tabucline [1pt]{1}
Corpus & 614 & 3.34 & 223 & 3.61 \\
Interact. & \textbf{1425} & \textbf{6.22} & 399 & \textbf{15.33} \\
\tabucline [1pt]{1}
\end{tabu}%
}
\caption{Number of unique dialogue states and average dialogue actions per state in the training corpus and in the RL interactions in two transfer learning setups.}
\label{tab:exploration}
\end{table}

\subsubsection{RL Learning Curve}\label{sssec:curve}
\begin{figure}[t]
    \centering
    \includegraphics[width=\linewidth]{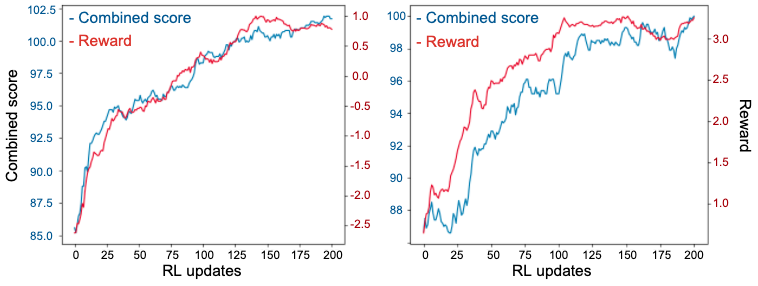}
    \caption{Learning curves observed on the dev set during \gls{rl} optimization. Two domain adaptation cases are presented, with restaurant (left) and hotel (right) as target domain respectively.}
    \label{fig:curve}
\end{figure}

Here we show that the designed reward structure is indeed a useful objective for training.
\fig{curve} shows learning curves of the model performance and the received (turn-level) rewards during \gls{rl} training.
The two examples are from the domain adaptation experiments in \ssectdot{transferlearning}, where restaurant (left) and hotel (right) are the target domain.
We can see that both the reward value and model performance are consistently improved during \gls{rl}, and their high correlation verifies the efficacy of the proposed reward design for training task-oriented dialogue systems.

\subsection{Human Evaluation}

The human assessment of dialogue quality is performed to confirm the improvements of the proposed methods.
400 dialogues, generated by the two trained agents, are evaluated by 14 human assessors.
Each assessor is shown a comparison of two dialogues where one dialogue is generated by the models using \gls{sl} and another is generated by the models after RL optimization.
Note that here we are evaluating the performance gain during interactions between two agents (\ssectdot{result_with_us}), instead of the gain in benchmark results by interacting with the static corpus (\ssectdot{benchmark}).
This is why the baseline is our \gls{sl} model instead of the existing state-of-the-art systems.

The assessor offers judgement regarding:
\begin{itemize} [itemsep=2pt,topsep=2pt] 
    \item Which dialogue system completes the task more successfully (\textbf{DS Success})?
    \item Which user simulator behaves more like a real human user (\textbf{US Human-like})?
    \item Which dialogue is more natural, fluent and efficient (\textbf{Dialogue Flow})?
\end{itemize}
The results with relative win ratio, shown in \tbl{human_eval}, are consistent with the automatic evaluation.
With the proposed RL optimisation, the \gls{ds} is more successful in dialogue completion.
More importantly, joint optimisation of the \gls{us} is found to produce more human-like behavior.
The improvement under the two agents leads to a more natural and efficient dialogue flow.

\begin{table}[t]
\centering
\resizebox{0.6\linewidth}{!}{%
\begin{tabu}{lcc}
\tabucline [1pt]{1}
\multicolumn{1}{c}{\textbf{Win Ratio (\%)}} & \textbf{SL} & \textbf{RL} \\ \tabucline [1pt]{1}
DS Success & 26.0 & \textbf{74.0} \\
US Human-like & 29.5 & \textbf{70.5} \\
Dialogue Flow & 21.0 & \textbf{79.0} \\
\tabucline [1pt]{1}
\end{tabu}%
}
\caption{Human assessment of the system quality under supervised learning and reinforcement learning.}
\label{tab:human_eval}
\end{table}
\section{Related Work}

In the emerging field of end-to-end \glspl{ds}, in which all components of a system are trained jointly \cite{Liu2017, wenN2N17,lei-etal-2018-sequicity}.
\gls{rl} methods have been used effectively to optimize end-to-end \glspl{ds} in \cite{dhingra-etal-2017-towards, liu2017e2e, zhao2019rethinking}, although using rule-based \glspl{us} or a fixed corpus for interaction.
Recent works utilise powerful transformers such as GPT-2 \cite{peng2020soloist,hosseini2020simple} or T5 \cite{lin2020mintl} for dialogue modeling and reach state-of-the-art performance; however, the area of having a user simulator involved during training is unexplored.
By comparison, this work uses a learned \gls{us} as the environment for \gls{rl}. The two agents we propose are able to generate abundant high-quality dialog examples and they can be extended easily to unseen domains. By utilizing an interactive environment instead of a fixed corpus, more dialogue strategies are explored and more dialogue states are visited.

There have been various approaches to building \glspl{us}.
In the research literature of \glspl{us},
one line of research is rule-based simulation such as the \gls{abus} \cite{Schatzmann2009TheHA, Li2016AUS}.
The \gls{abus}'s structure is such that it has to be re-designed for different tasks, which presents challenges in shifting to new scenarios.
Another line of work is data-driven modelling.
\citet{Asri2016ASM} modelled user simulation as a seq2seq task, where the output is a sequence of user dialogue acts the level of semantics.
\citet{Gur2018UserMF} proposed a variational hierarchical seq2seq framework to introduce more diversity in generating the user dialogue act. \citet{Kreyssig2018NeuralUS} introduced the \gls{nus}, a seq2seq model that learns the user behaviour entirely from a corpus, generates natural language instead of dialogue acts and possesses an explicit goal representation. The \gls{nus} outperformed the \gls{abus} on several metrics. \citet{Kreyssig2018MEng} also compared the \gls{nus} and \gls{abus} to a combination of the \gls{abus} with an NLG component.
However, none of these prior works are suitable for modelling complex, multi-domain dialogues in an end-to-end fashion.
By contrast, the user model proposed here consumes and generates text and so can be directly employed to interact with the \gls{ds}, communicating via natural language.

The literature on joint optimization of the \gls{ds} and the \gls{us} is line of research most relevant to our work.
\citet{takanobu2020multi} proposed a hybrid value network using MARL \cite{Lowe2017MultiAgentAF} with role-aware reward decomposition used in optimising the dialogue manager. However, their model requires separate NLU/NLG models to interact via natural language, which hinders its application in the transfer learning to new domains.
\citet{Liu2017IterativePL,papangelis-etal-2019-collaborative} learn both the \gls{ds} and the \gls{us} in a (partially) end-to-end manner.
However, their systems are designed for the single-domain dataset (DSTC2) and cannot handle the complexity of multi-domain dialogues: 1) their models can only predict one dialogue act per turn, which is not sophisticated enough for modelling multiple concurrent dialogue acts;
2) the simple DST components cannot achieve satisfactory performance in the multi-domain setup;
3) the user goal change is not modelled along the dialogue proceeds, which we found in our experiments very important for learning complex behaviors of user simulators.
Relative to these three publications, this paper focuses on joint training of two fully end-to-end agents that are able to participate in complex multi-domain dialogues. More importantly, it is shown that the proposed framework is highly effective for transfer learning, which is a novel contribution relative to previous work.
\section{Conclusion and Future Work}


We propose a novel joint learning framework of training both the \gls{ds} and the \gls{us} for complex multi-domain dialogues.
Under the low-resource scenarios, the two agents can generate more dialogue data through interacting with each other and their behaviors can be significantly improved using RL through this self-play strategy.
Two types of reward are investigated and the turn-level reward benefits more due to its fine-grained structure.
Experiments shows that our framework outperforms previously published results on the MultiWOZ dataset.
In two transfer learning setups, our method can further improves the well-performed EWC models and bootstraps the final performance largely.
Future work will focus on improving the two agents' underlying capability with the powerful transformer-based models.

\section*{Acknowledgements}
Bo-Hsiang Tseng is supported by Cambridge Trust and the Ministry of Education, Taiwan.
Florian Kreyssig is funded by an EPSRC Doctoral Training Partnership Award.
This work has been performed using resources provided by the Cambridge Tier-2 system operated by the University of Cambridge Research Computing Service (http://www.hpc.cam.ac.uk) funded by EPSRC Tier-2 capital grant EP/P020259/1.

\newpage
\clearpage

\bibliographystyle{acl_natbib}
\bibliography{anthology,acl2021}

\clearpage
\appendix
\section{Appendices}

\subsection{Training Details} \label{ssec:config}
Both the \gls{ds} and the \gls{us} are trained in an end-to-end fashion using the Adam optimizer. 
The sizes of the embedding and of the hidden layers are set to 300. During supervised training, the batch size is 100 and the learning rate is 0.001, while during \gls{rl}, 10 is used as the batch size and 0.0001 as the learning rate for stability. We set the discounting factor $\gamma$ to 1.
The computing infrastructure used is Linux 4.4.0-138-generic x86\_64 with the NVIDIA GPU GTX-1080. Average run time per model using 100\% training data is around 6 hours. Model parameters is around 11M in total.

The turn-level rewards used for the best models in benchmark results are reported in \tbl{reward} below. All rewards are tuned based on the combined score of the validation performance averaged over three seeds. As for dialogue-level rewards, a positive reward $1.0$ will be given if a dialogue is successful.

\begin{table}[h!]
\centering
\resizebox{0.9\linewidth}{!}{%
\begin{tabu}{lcccccc}
\tabucline [1pt]{1}
\multicolumn{1}{c}{\multirow{2}{*}{Model}} & \multicolumn{3}{c}{\textbf{Rewards on DS}} & \multicolumn{3}{c}{\textbf{Rewards on US}} \\ \cline{2-7} 
\multicolumn{1}{c}{} & $r^{req}$ & $r^{pro}$ & $r^{ans}$ & $r^{req}$ & $r^{inf}$ & $r^{ans}$ \\ \tabucline [1pt]{1}
RL-DS & 0, -1 & 0, -5 & 2.5, -5 & 0, 0 & 0, 0 & 0, 0 \\
RL-Joint & 0, -1 & 0, -5 & 2.5, -5 & 0, -1 & 1, -1 & 1, -1 \\
\tabucline [1pt]{1}
\end{tabu}%
}
\caption{The configuration of turn-level rewards for each best model in the reported benchmark results. Each reward has positive and negative values.}
\label{tab:reward}
\vspace{-5mm}
\end{table}

\subsection{Details of Dataset}
As noted in the paper, we follow the original split of the MultiWOZ dataset and the number of dialogues for train/dev/test split is 8420/1000/1000. Data statistics of the number of dialogues in the two transfer learning scenarios are provided in \twotbl{domain-data}{s2m-data}.


\begin{table}[ht]
\centering
\resizebox{0.9\linewidth}{!}{%
\begin{tabu}{lccccc}
\tabucline [1pt]{1}
\textbf{Data} & \multicolumn{1}{l}{\textbf{Restaurant}} & \multicolumn{1}{l}{\textbf{Hotel}} & \multicolumn{1}{l}{\textbf{Attraction}} & \multicolumn{1}{l}{\textbf{Train}} & \multicolumn{1}{l}{\textbf{Taxi}} \\ \tabucline [1pt]{1}
Train & 300 & 300 & 300 & 300 & 300 \\
Dev & 438 & 415 & 400 & 484 & 206 \\
Test & 437 & 394 & 396 & 495 & 195 \\
\tabucline [1pt]{1}
\end{tabu}%
}
\caption{Number of dialogues of the splits in each domain adaptation}
\label{tab:domain-data}
\vspace{-2mm}
\end{table}

\begin{table}[ht]
\centering
\resizebox{0.9\linewidth}{!}{%
\begin{tabu}{lrrrrrr}
\tabucline [1pt]{1}
\textbf{Data} & \multicolumn{1}{l}{\textbf{H+T}} & \multicolumn{1}{l}{\textbf{R+T}} & \multicolumn{1}{l}{\textbf{A+T}} & \multicolumn{1}{l}{\textbf{A+H+X}} & \multicolumn{1}{l}{\textbf{H+R+X}} & \multicolumn{1}{l}{\textbf{A+R+X}} \\ \tabucline [1pt]{1}
Train & 100 & 100 & 100 & 100 & 100 & 100 \\
Dev & 149 & 157 & 148 & 110 & 100 & 131 \\
Test & 144 & 155 & 163 & 92 & 91 & 129 \\
\tabucline [1pt]{1}
\end{tabu}%
}
\caption{Number of dialogues of the splits in each scenario in single-to-multi domain transfer learning.}
\label{tab:s2m-data}
\vspace{-2mm}
\end{table}

\subsection{Error Analysis}
The error analysis of each domain adaptation cases are provided in \twotbl{error_ds}{error_us}.

\begin{table}[h!]
\centering
\resizebox{1\linewidth}{!}{%
\begin{tabu}{lccccc||c}
\tabucline [1pt]{1}
\multicolumn{1}{c}{Model} & \textbf{Restaurant} & \textbf{Hotel} & \textbf{Attraction} & \textbf{Train} & \textbf{Taxi} & \textbf{Average} \\ \tabucline [1pt]{1}
\multicolumn{7}{c}{Missing provision rate (\%)} \\ \hline
SL & 14.84 & 24.74 & 29.08 & 8.61 & 10.68 & 17.59 \\
RL & \textbf{1.45} & \textbf{2.33} & \textbf{7.83} & 0.41 & 1.62 & \textbf{2.73} \\ \tabucline [1pt]{1}
\multicolumn{7}{c}{Missing answer rate (\%)} \\ \hline
SL & 44.39 & 61.52 & 32.33 & 11.95 & 34.78 & 36.99 \\
RL & \textbf{5.64} & 20.04 & 10.35 & 2.58 & \textbf{9.09} & \textbf{9.54} \\
\tabucline [1pt]{1}
\end{tabu}%
}
\caption{Error analysis on dialogue system on each domain in terms of two behaviors. Lower the better.}
\label{tab:error_ds}
\vspace{-2mm}
\end{table}

\begin{table}[h!]
\centering
\resizebox{1\linewidth}{!}{%
\begin{tabu}{lccccc||c}
\tabucline [1pt]{1}
\multicolumn{1}{c}{Model} & \textbf{Restaurant} & \textbf{Hotel} & \textbf{Attraction} & \textbf{Train} & \textbf{Taxi} & \textbf{Average} \\ \tabucline [1pt]{1}
\multicolumn{7}{c}{Repeat inform rate (\%)} \\ \hline
SL & 8.45 & 9.15 & 16.81 & 9.69 & 6.48 & 10.12 \\
RL & \textbf{1.84} & \textbf{1.28} & \textbf{0.70} & \textbf{1.47} & \textbf{2.05} & \textbf{1.47} \\ \tabucline [1pt]{1}
\multicolumn{7}{c}{Missing answer rate (\%)} \\ \hline
SL & 27.84 & 56.41 & 64.68 & 35.41 & 52.00 & 47.27 \\
RL & \textbf{24.44} & \textbf{40.74} & \textbf{22.95} & \textbf{26.44} & \textbf{48.42} & \textbf{32.60} \\
\tabucline [1pt]{1}
\end{tabu}%
}
\caption{Error analysis on user simulator on each domain in terms of two behaviors. Lower the better.}
\label{tab:error_us}
\vspace{-5mm}
\end{table}

\subsection{Exploration} 
The detailed numbers of explored dialogue states and the average of unique dialogue actions per state in each case of two transfer learning scenarios are provided in \twotbl{explore-domains}{explore-s2ms}.

\begin{table}[h!]
\centering
\resizebox{\linewidth}{!}{%
\begin{tabu}{lccccc||c}
\tabucline [1pt]{1}
\multicolumn{1}{c}{\textbf{Model}} & \textbf{Restaurant} & \textbf{Hotel} & \textbf{Attraction} & \textbf{Train} & \textbf{Taxi} & \textbf{Average} \\ \tabucline [1pt]{1}
\multicolumn{7}{c}{Number of dialogue states} \\ \hline
SL & 514 & 726 & 545 & 513 & 774 & 614 \\
RL & \textbf{1458} & \textbf{1666} & 1087 & \textbf{601} & 2313 & \textbf{1425} \\ \tabucline [1pt]{1}
\multicolumn{7}{c}{Average of dialogue actions per state} \\ \hline
SL & 3.51 & 3.10 & 3.74 & 3.70 & 2.65 & 3.34 \\
RL & 4.92 & \textbf{5.49} & \textbf{7.61} & \textbf{7.98} & 5.10 & \textbf{6.22} \\
\tabucline [1pt]{1}
\end{tabu}%
}
\caption{Number of dialogue states and average of dialogue actions per state in each domain adaptation case.}
\label{tab:explore-domains}
\vspace{-5mm}
\end{table}

\begin{table}[h!]
\centering
\resizebox{\linewidth}{!}{%
\begin{tabu}{lcccccc||c}
\tabucline [1pt]{1}
\multicolumn{1}{c}{\textbf{Model}} & \textbf{H+T} & \textbf{R+T} & \textbf{A+T} & \textbf{A+H+X} & \textbf{H+R+X} & \textbf{A+R+X} & \textbf{Average} \\ \tabucline [1pt]{1}
\multicolumn{8}{c}{Number of dialogue states} \\ \hline
SL & 250 & 184 & 118 & 263 & 352 & 172 & 223 \\
RL & 523 & 294 & \textbf{208} & 348 & \textbf{636} & \textbf{383} & 399 \\ \tabucline [1pt]{1}
\multicolumn{8}{c}{Average of dialogue actions} \\ \hline
SL & 2.84 & 3.95 & 5.40 & 3.00 & 2.43 & 4.05 & 3.61 \\
RL & 6.98 & \textbf{13.56} & \textbf{17.90} & 21.03 & \textbf{11.30} & \textbf{21.20} & \textbf{15.33} \\
\tabucline [1pt]{1}
\end{tabu}%
}
\caption{Number of dialogue states and average of dialogue actions per state in each single-to-multi domain case.}
\label{tab:explore-s2ms}
\end{table}

\newpage
\clearpage

\subsection{Transfer Learning}
Here we provide the results in success rate in two transfer learning setups.

\begin{table}[h]
\centering
\resizebox{\linewidth}{!}{%
\begin{tabu}{lccccc||c}
\tabucline [1pt]{1}
\textbf{Model} & \textbf{Restaurant} & \textbf{Hotel} & \textbf{Attraction} & \textbf{Train} & \textbf{Taxi} & \textbf{Avg.} \\ \hline
\multicolumn{7}{c}{Belief State = Predicted} \\ \hline
Source & 5.0 &	10.9 &	5.4 &	36.2 &	0.0 &	11.5 \\ \hdashline
Naive & 26.4 &	35.8 &	41.0 &	48.0 &	35.0 &	37.2 \\
EWC & 35.9 &	37.8 &	47.6 &	47.7 &	55.2 &	44.9 \\ \hdashline
Naive+RL & 36.8 &	46.0 &	46.2 &	\textbf{49.8} &	41.4 &	44.1 \\
EWC+RL & \textbf{42.3} & \textbf{47.7} & \textbf{51.9} &	48.5 &	\textbf{63.9} &	\textbf{50.8} \\ \tabucline [1pt]{1}
\multicolumn{7}{c}{Belief State = Oracle} \\ \hline
Source & 11.8 & 18.6 &	9.1 &	45.3 &	0.0 &	17.0 \\ \hdashline
Naive & 60.7 &	62.1 &	46.8 &	73.9 &	67.2 &	62.1 \\
EWC & 59.3 &	62.4 &	64.5 &	79.5 &	74.5 &	68.0 \\ \hdashline
Naive+RL & 73.1 &	76.2 &	58.5 &	82.6 &	81.4 &	74.4 \\
EWC+RL & \textbf{73.3} &	\textbf{79.3} &	\textbf{70.5} &	\textbf{82.8} &	\textbf{84.6} &	\textbf{78.1} \\
\tabucline [1pt]{1}
\end{tabu}%
}
\caption{Success rate in domain adaptation. 300 dialogues are used for each target domain adaptation.}
\label{tab:domain_transfer_succ}
\end{table}

\begin{table}[h]
\centering
\resizebox{\linewidth}{!}{%
\begin{tabu}{lcccccc||c}
\tabucline [1pt]{1}
\textbf{Model} & \multicolumn{1}{l}{\textbf{H+T}} & \multicolumn{1}{l}{\textbf{R+T}} & \multicolumn{1}{l}{\textbf{A+T}} & \multicolumn{1}{l}{\textbf{A+H+X}} & \multicolumn{1}{l}{\textbf{H+R+X}} & \multicolumn{1}{l}{\textbf{A+R+X}} & \textbf{Avg.} \\ \tabucline [1pt]{1}
\multicolumn{8}{c}{Belief State = Predicted} \\ \hline
Source & 30.8 &	40.7 &	15.1 &	5.1 &	8.8 & 4.9 & 17.6 \\ \hdashline
Naive & 40.7 & 48.8 & 37.8 & 16.3 & 15.8 & 19.9 & 29.9 \\
EWC & 41.0 & 50.5 & 42.3 & 19.6 & 17.2 & 20.4 & 31.8 \\ \hdashline
Naive+RL & \textbf{47.2} & 53.8 & 42.9 & 18.1 & \textbf{22.0} & 24.0 & 34.7 \\
EWC+RL & 45.8 & \textbf{57.9} & \textbf{44.6} & \textbf{20.7} & 20.2 & \textbf{24.6} & \textbf{35.6} \\ \tabucline [1pt]{1}
\multicolumn{8}{c}{Belief State = Oracle} \\ \hline
Source & 60.4 & 74.6 & 41.3 & 13.4 & 28.2 & 19.9 & 39.6 \\ \hdashline
Naive & 67.1 & 76.6 & 54.4 & 48.2 & 53.9 & 46.3 & 57.7 \\
EWC & 74.5 & 78.1 & 60.3 & 43.8 & 57.9 & 50.4 & 60.8 \\ \hdashline
Naive+RL & \textbf{79.6} & 83.4 & 60.3 & \textbf{57.3} & 68.1 & 60.2 & 68.2 \\
EWC+RL & 79.2 & \textbf{84.5} & \textbf{66.1} & 55.4 & \textbf{70.0} & \textbf{61.5} & \textbf{69.4} \\
\tabucline [1pt]{1}
\end{tabu}%
}
\caption{Success rate in single-to-multiple domain transfer where 100 dialogues on each target scenario are used for adaptation. R, H, A, T, X represent Restaurant, Hotel, Attraction, Train, Taxi domain.}
\label{tab:single-to-multi_succ}
\end{table}

\subsection{Generated Dialogue Examples}
Here we provides two dialogue pairs examples that are generated by the two agents' interaction using supervised learning (SL) and using RL respectively in the following pages.
As seen in the \tbl{example1} first dialogue, since the SL user model answers wrong to the system's request, the system keeps asking the unsolved question, leading to an unsuccessful dialogue with loop (highlighted in color).
On the other hand, the interaction between RL models (second dialogue) is much smoother and efficient since both agents answer correctly to each other. This indicates the designed rewards are useful to improve the agents' behaviors.
Similar trend can be observed in \tbl{example2}, where the user model keeps stating the same information about restaurant.

\newpage
\clearpage

\begin{table*}[ht]
\centering
\resizebox{0.925\linewidth}{!}{%
\begin{tabu}{l}
\tabucline [1pt]{1}
\multicolumn{1}{c}{\begin{tabular}[c]{@{}c@{}}\textbf{User Goal}\\
\textbf{Train}: destination: cambridge; day: wednesday; arriveBy: 18:15; \\ departure: kings lynn; request={[}trainID, price, leaveAt{]}\\
\textbf{Attraction}: name: club salsa; request={[}address{]}\\
\end{tabular}}\\ \tabucline [1pt]{1}
\multicolumn{1}{c}{Dialogue generated by two agents trained using SL} \\ \multicolumn{1}{c}{\textcolor{MyDarkRed}{(Dialogue has a loop and fails as the provided train is incorrect due to the wrong departure)}} \\
\hline \hline
00-usr: i am looking for a train that leaves on wednesday and arrives at cambridge . \\
01-sys: there are few trains that meet your criteria . what time would you like to leave ? \\
02-usr: i would like to arrive by 18:15 . \\
03-sys: there are few trains leaving london kings cross to cambridge on wednesday . \\ \hspace{1.2cm}  \textcolor{MyDarkOrange}{what time would you like to leave ?} \\
04-usr: \textcolor{MyGold}{i would like to arrive by 18:15 .} \\
05-sys: there are few trains leaving london kings cross to cambridge . \\ \hspace{1.2cm}  \textcolor{MyDarkOrange}{is there a particular time you would like to leave ?} \\
06-usr: \textcolor{MyGold}{i would like to arrive by 18:15 .} \\
07-sys: there are few trains leaving london kings cross to cambridge . \\ \hspace{1.2cm}  \textcolor{MyDarkOrange}{is there a particular time you would like to leave ?} \\
08-usr: \textcolor{MyGold}{i would like to arrive by 18:15 .} \\
09-sys: there are few trains leaving london kings cross to cambridge . \\ \hspace{1.2cm}  \textcolor{MyDarkOrange}{is there a particular time you would like to leave ?} \\
10-usr: \textcolor{MyGold}{i would like to arrive by 18:15 .} \\
11-sys: there are few trains leaving london kings cross to cambridge . \textcolor{MyDarkOrange}{what time would you like to leave ?} \\
12-usr: \textcolor{MyGold}{i would like to arrive by 18:15 .} \\
13-sys: there are few trains leaving london kings cross to cambridge . \textcolor{MyDarkOrange}{what time would you like to leave ?} \\
14-usr: \textcolor{MyGold}{i would like to arrive by 18:15 .} \\
15-sys: there are few trains leaving london kings cross to cambridge . \textcolor{MyDarkOrange}{what time would you like to leave ?} \\
16-usr: \textcolor{MyGold}{i would like to arrive by 18:15 .} \\
17-sys: there are few trains leaving london kings cross to cambridge . \textcolor{MyDarkOrange}{what time would you like to leave ?} \\
18-usr: \textcolor{MyGold}{i would like to arrive by 18:15 .} \\
19-sys: TR0269 arrives at 18:15 . would you like me to book it for you ? \\
20-usr: yes , please . i need the price and travel time . \\
21-sys: the train id is TR0269 , the price is 10 and the travel time is 50 . \\
22-usr: i am also looking for a particular attraction called club salsa . \\
23-sys: club salsa is a nightclub in the centre . the address is 1 station road . entrance is free . \\ \hspace{1.2cm}  is there anything else i can help you with ? \\
24-usr: that is all i need . thank you for your help . \\
25-sys: thank you for using our system ! \\

\tabucline [1pt]{1} \tabucline [1pt]{1}
\multicolumn{1}{c}{Dialogue generated by two agents trained using RL} \\ \multicolumn{1}{c}{\textcolor{MyDarkGreen}{(Dialogue is smooth and successful)}} \\ 
\hline \hline
00-usr: i am looking for a train from kings lynn on wednesday . \\
01-sys: what is your destination ? \\
02-usr: i need to arrive in cambridge by 18:15 . \\
03-sys: i have train TR0415 that leaves at 17:11 and arrives at 17:58 . would you like me to book it for you ? \\
04-usr: yes , please book 2 tickets . \\
05-sys: booking was successful , the total fee is 9.80 pounds payable at the station . \\ \hspace{1.2cm} reference number is fztwszhh . is there anything else i can help you with ? \\
06-usr: yes , i am also looking for a particular attraction called club salsa . \\
07-sys: club salsa is located at 1 station road , postcode cb12jb . is there anything else i can help you with ? \\
08-usr: no , that s all i need today . thank you . \\
09-sys: thank you for using our service . have a great day ! \\
\tabucline [1pt]{1}
\end{tabu}%
}
\caption{Example 1: two dialogues generated by SL system and RL system respectively. The dialogue loop with repeated information is highlighted. Output utterances are lexicalised with corresponding slot values (belief state for system; user goal for user). If the system generates attributes that has not existing in the belief state, the values are randomly sampled from the ontology, e.g., the train departure "london kings cross" in the first dialogue.}
\label{tab:example1}
\end{table*}

\begin{table*}[ht]
\centering
\resizebox{0.925\linewidth}{!}{%
\begin{tabu}{l}
\tabucline [1pt]{1}
\multicolumn{1}{c}{\begin{tabular}[c]{@{}c@{}}\textbf{User Goal}\\
\textbf{Restaurant}: pricerange=expensive; area=west; request={[}phone, postcode{]}\\
\textbf{Train}: destination: cambridge; day: friday; arriveBy: 21:30; \\ departure: norwich; request={[}price, leaveAt{]}\\
\end{tabular}}\\ \tabucline [1pt]{1}
\multicolumn{1}{c}{Dialogue generated by two agents trained using SL} \\ \multicolumn{1}{c}{\textcolor{MyDarkRed}{(Dialogue has a loop and fails as the involved restaurant is cheap instead of expensive)}} \\
\hline \hline
00-usr: i am looking for a expensive restaurant in the west . \\
01-sys: there are few restaurant -s that meet your criteria . what type of food would you like ? \\
02-usr: i would like to book a table for 1 at 1 . can i get the phone number for thanh binh ? \\
03-sys: thanh binh phone number is 01223362456 . \\
04-usr: \textcolor{MyDarkOrange}{i am also looking for a restaurant called thanh binh .} \\
05-sys: \textcolor{MyGold}{thanh binh is located at 17 Magdalene Street City Centre , phone number is 01223362456 .} \\
06-usr: \textcolor{MyDarkOrange}{i am also looking for a restaurant called thanh binh .} \\
07-sys: \textcolor{MyGold}{thanh binh is located at 17 Magdalene Street City Centre , phone number is 01223362456 .} \\
08-usr: \textcolor{MyDarkOrange}{i am also looking for a restaurant called thanh binh .} \\
09-sys: \textcolor{MyGold}{thanh binh is located at 17 Magdalene Street City Centre , phone number is 01223362456 .} \\
10-usr: \textcolor{MyDarkOrange}{i am also looking for a restaurant called thanh binh .} \\
11-sys: \textcolor{MyGold}{thanh binh is located at 17 Magdalene Street City Centre , phone number is 01223362456 .} \\
12-usr: thank you . \textcolor{MyDarkOrange}{i am also looking for a restaurant called thanh binh .} \\
13-sys: \textcolor{MyGold}{thanh binh is located at 17 Magdalene Street City Centre , phone number is 01223362456 .} \\
14-usr: can you give me the postcode for thanh binh ? \\
15-sys: the postcode is cb21ab . is there anything else i can help you with ? \\
16-usr: i am also looking for a train from norwich to cambridge on friday . \\
17-sys: there are 1 trains that fit your criteria . what time would you like to leave or arrive by ? \\
18-usr: i need to arrive by 21:30 . \\
19-sys: None arrives at 21:30 . would you like me to book it for you ? \\
20-usr: no , but can you give me the departure time and departure time ? \\
21-sys: the departure time is 05:16 . \\
22-usr: thank you so much . that s all i need today . \\
23-sys: thank you for using our service . have a great day ! \\
\tabucline [1pt]{1} \tabucline [1pt]{1}
\multicolumn{1}{c}{Dialogue generated by two agents trained using RL} \\ \multicolumn{1}{c}{\textcolor{MyDarkGreen}{(Dialogue is smooth and successful)}} \\ 
\hline \hline
00-usr: i am looking for a expensive restaurant in the west . \\
01-sys: tandoori palace is a expensive indian restaurant in the west . \\
02-usr: what is the phone number and postcode ? \\
03-sys: the phone number is 01223506055 and the postcode is cb43le . \\
04-usr: thank you . i also need a train from norwich to cambridge on friday . \\
05-sys: i can help you with that . what time would you like to leave or arrive by ? \\
06-usr: i need to arrive by 21:30 . \\
07-sys: TR8009 arrives at 21:30 . \\
08-usr: what is the travel time and price ? \\
09-sys: the travel time is 79 minutes and the price is 17.60 pounds . \\
10-usr: what is the departure time ? \\
11-sys: the train leaves at 18:16 . \\
12-usr: what is the train id ? \\
13-sys: the train id is TR8009 . \\
14-usr: thank you very much . \\
15-sys: thank you , goodbye . \\
\tabucline [1pt]{1}
\end{tabu}%
}
\caption{Example 2: two dialogues generated by SL system and RL system respectively. The dialogue loop with repeated information is highlighted. Output utterances are lexicalised with corresponding slot values (belief state for system; user goal for user). If the user generates attributes beyond the given user goal, the values are randomly sampled from the ontology, e.g., the restaurant name "thanh binh" in the first dialogue.}
\label{tab:example2}
\end{table*}

\end{document}